\documentclass[subscriptcorrection,upint,varvw,barcolor=Goldenrod3,mathalfa=cal=euler,balance,hyphenate,french,pdf-a,nofoot,nolists]{asmejour} %

\usepackage{booktabs}
\usepackage{graphicx}
\usepackage{colortbl}

\usepackage{mdframed}
\usepackage{xcolor}

\definecolor{softgreen}{RGB}{118, 158, 118} 

\newmdenv[
  topline=false, 
  bottomline=false,
  rightline=false,
  leftline=false,
  linewidth=2pt,
  linecolor=softgreen,
  backgroundcolor=gray!20,
  leftmargin=10pt,
  rightmargin=10pt,
  innertopmargin=10pt,
  innerbottommargin=10pt,
  font=\ttfamily,
  frametitlebackgroundcolor=softgreen, 
  frametitlerule=false, 
  frametitlerulewidth=0pt,
  frametitlefont=\bfseries\color{white}, 
]{customquote}

\hypersetup{%
	pdfauthor={Daniele Grandi},                       		   	
	pdftitle={JCISE LLM - LLMs Evaluating LLMs for Material Selection},                  	
	pdfkeywords={},
	pdfsubject = {},			
}

\JourName{Computing and Information Science in Engineering}

\begin{document}


\SetAuthorBlock{Daniele Grandi\CorrespondingAuthor}{Autodesk Research,\\
   San Francisco, CA, 94111\\
   daniele.grandi@autodesk.com} 


\SetAuthorBlock{Yash Patawari Jain}{%
   Carnegie Mellon University,\\
   Pittsburgh, PA 15289\\
    ypatawar@andrew.cmu.edu
}

\SetAuthorBlock{Allin Groom}{%
   Autodesk Research,\\
   London, UK \\
    allin.groom@autodesk.com
}

\SetAuthorBlock{Brandon Cramer}{%
   Autodesk Research,\\
   Boston, MA \\
    brandon.stewart.cramer@autodesk.com
}

\SetAuthorBlock{Christopher McComb}{Department of Mechanical Engineering,\\
   Carnegie Mellon University,\\
   Pittsburgh, PA 15289\\
   ccm@cmu.edu} 

\title{Evaluating Large Language Models for Material Selection}


\begin{abstract}
Material selection is a crucial step in conceptual design due to its significant impact on the functionality, aesthetics, manufacturability, and sustainability impact of the final product. This study investigates the use of Large Language Models (LLMs) for material selection in the product design process and compares the performance of LLMs against expert choices for various design scenarios. By collecting a dataset of expert material preferences, the study provides a basis for evaluating how well LLMs can align with expert recommendations through prompt engineering and hyperparameter tuning.
The divergence between LLM and expert recommendations is measured across different model configurations, prompt strategies, and temperature settings. This approach allows for a detailed analysis of factors influencing the LLMs' effectiveness in recommending materials.
The results from this study highlight two failure modes, and identify parallel prompting as a useful prompt-engineering method when using LLMs for material selection. The findings further suggest that, while LLMs can provide valuable assistance, their recommendations often vary significantly from those of human experts. This discrepancy underscores the need for further research into how LLMs can be better tailored to replicate expert decision-making in material selection.
This work contributes to the growing body of knowledge on how LLMs can be integrated into the design process, offering insights into their current limitations and potential for future improvements.
\end{abstract}

\date{}

\maketitle 


\begin{figure*}[h]
    \centering
    \includegraphics[width=\textwidth]{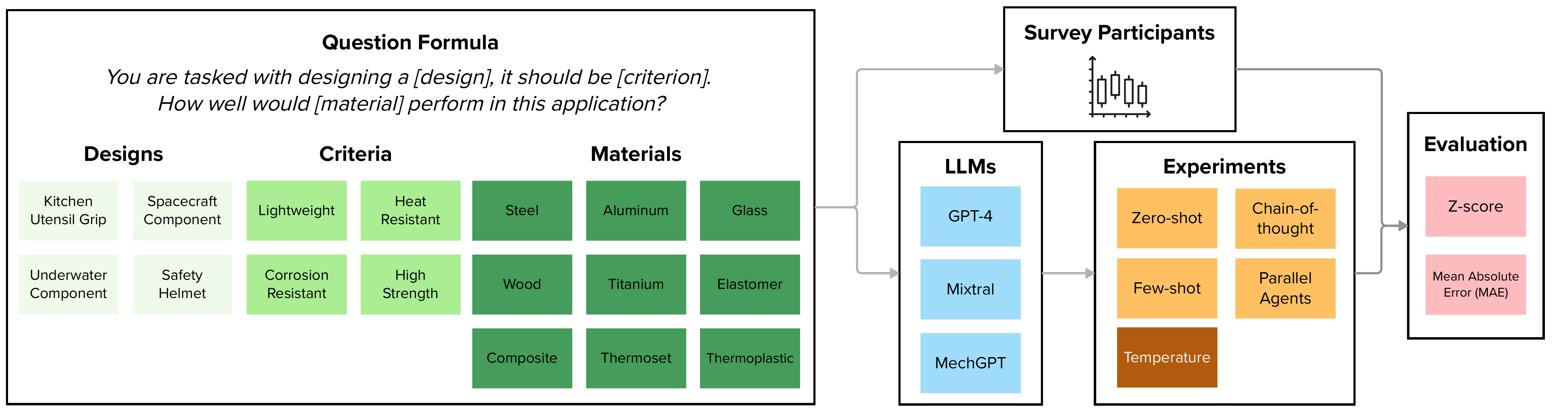}
    \caption{Overview of the method used to create the corpus of questions submitted to survey participants and to the LLMs, the experiments used to evaluate the LLMs, and the evaluation metrics used to compare the LLM results to the survey responses.}
    \label{fig:overview}
\end{figure*}

\section{Introduction}


Material selection plays an important role in the conceptual design phase of engineering projects. The choice of materials directly impacts the overall performance, durability, and efficiency of a product. Selecting the right material ensures that the design meets its intended purpose, withstands loads, and performs optimally \cite{ashby_selection_2004}. Material selection affects manufacturing costs, maintenance expenses, and the product’s lifecycle cost. Opting for cost-effective materials without compromising quality is essential \cite{giachetti_manufacturing_1997}. For instance, using lightweight materials can reduce transportation costs and energy consumption. In critical applications, material failure can have catastrophic consequences \cite{noauthor_what_nodate}. Proper material selection ensures safety, reliability, and longevity. For instance, selecting materials with corrosion-resistant properties for marine structures can prevent premature deterioration. As an added complexity, there is also an emerging requirement for the sustainability of materials to be considered during selection, specifically how choices impact the environment. Designers must consider factors like recyclability, energy consumption during production, and the broader carbon footprint associated with a candidate material. Materials influence the ease of manufacturing and assembly \cite{callister2007materials}. Some materials are easier to shape, weld, or join, leading to efficient production processes. Conversely, poor material choices can complicate manufacturing and increase costs. Material aesthetics influence the product’s visual appeal and user experience. Whether it’s a sleek smartphone casing or a comfortable chair, material selection affects how users perceive and interact with the design.

When treated as an optimization problem, material selection involves optimizing both the underlying geometry and selecting an appropriate material. While some problems allow for separate optimization of material and geometry, in general, they are intertwined. The discrete nature of material selection poses challenges when combined with gradient-based geometry optimization. Researchers have proposed innovative approaches to address this complex problem.
One such approach involves using variational autoencoders (VAEs). This framework has been demonstrated effectively in truss design scenarios, where optimal materials are chosen from a database while simultaneously optimizing cross-sectional areas \cite{chandrasekhar_integrating_2022}. However, material selection as an optimization problem has its limitations. The multi-objective nature of material selection involves conflicting criteria such as strength, weight, cost, and environmental impact. Balancing these objectives can be challenging, especially when considering multiple parts or components in a system \cite{aires_new_2022}. Material selection inherently involves discrete choices from a finite set of options. Integrating discrete material selection with continuous geometry optimization methods can be computationally demanding and may require specialized algorithms \cite{chandrasekhar_integrating_2022}. Material properties can exhibit variability due to manufacturing processes, environmental conditions, and other factors. Incorporating uncertainty into optimization models is essential but adds complexity \cite{ermolaeva_materials_2002}. Selecting an optimal material often involves trade-offs. For instance, choosing a lightweight material may sacrifice strength. These trade-offs require careful consideration during the optimization process, and might result in multiple right answers to varying degrees \cite{ashby_selection_2004}. Because of that, and given that material requirements are often textual, it is possible that Large Language Models (LLMs) could support the material selection task, similarly to how they have been used for conceptual design inspiration~\cite{ma_conceptual_2023}. 

In recent years, LLMs, or large-scale pre-trained neural networks, have emerged as powerful tools for natural language understanding. These models can process and generate human-like text, making them well-suited for handling textual material requirements. But can LLMs help designers navigate the multifaceted material selection landscape?
In this study, to capture designer's preferences towards certain material categories in different design scenarios, we collect a dataset of material preference from experts, across a set of design cases and design criteria. 
We then conduct a series of experiments to evaluate whether model choice, prompt engineering techniques, or hyperparameter tuning can help guide the LLMs toward more expert recommendations for a material selection task. 

We focus on two primary evaluation metrics: the z-score and the mean absolute error. These metrics serve as quantitative measures to evaluate the performance of materials against various criteria. The z-score quantifies how many standard deviations a material property deviates from the mean of a reference dataset. The mean absolute error (or Manhattan distance) measures the absolute difference between the corresponding components of the two data sets.

This paper is guided by the following research questions:
\begin{enumerate}
    \item 
    To what extent do LLMs exhibit bias toward specific materials?
    \item What effective methods can be employed to steer the LLMs towards appropriate material selection patterns?
\end{enumerate}

To support further development of LLM methods and evaluation on our data, the survey data, generated data, and code is available on Github\footnote{\href{https://github.com/grndnl/llm_material_selection_jcise}{https://github.com/grndnl/llm_material_selection_jcise/releases/tag/v1.0}}.

\section{Background}
\label{sec:background}

In this section we review works related to material selection and associated challenges, large language models and their application to design, and motivation specific to this problem.

\subsection{Challenges in Material Selection}
Material selection is a critical part of designing and producing any physical object. Material selection occurs in the early stages of the design workflow and maintains relevance beyond the useful life of a product. Materials directly influence the functionality, aesthetics, economic viability, manufacturing feasibility, and ultimately its environmental impact of a design \cite{zarandi_material_2011,bi_energy-aware_2017,noauthor_materials_nodate}. 
M. F. Ashby is often cited for presenting a systematic approach to material selection through the use of bubble plots, known as "Ashby" diagrams, which allow a designer to evaluate up to two material properties to identify those materials that perform above a desired threshold \cite{ashby_selection_2004}. This approach requires an intimate understanding of a product's design intent, the design priorities (such as low mass), constraints (manufacturing process), and other requirements relevant to the object being designed (industry regulations).
In recent years, additional factors have also become increasingly important to consider. Sustainability, for example, is a growing global concern, and manufacturing alone is reported to contribute significantly to resource consumption and greenhouse gas emissions \cite{us_epa_inventory_2024}. Thus, selecting materials with lower environmental impact, such as recycled content or those requiring less energy to produce, aligns with ethical practices and growing consumer expectations \cite{kishita_checklist-based_2010,banu_joint_2024,ramani_integrated_2010}. Material availability is also becoming a critical consideration due to supply chain disruptions, geopolitical challenges, or regulations on material use. The growing complexity of design requirements, does not reduce the implications of improper material selection which can lead to increased overall costs, product failure, or greater environmental harm \cite{albinana_framework_2012}. 

In product design, material selection can be broken down into a general five-step procedure: (1) establishing design requirements, (2) screening materials, (3) ranking materials, (4) researching material candidates, and (5) applying constraints to the selection process \cite{bhat_aerospace_2018}. Performance indices and material property charts, called Ashby diagrams, are often used to visualize, filter, and cluster materials \cite{bhat_aerospace_2018,doi:10.1179/mst.1989.5.6.517}. 

Traditionally, material selection has relied heavily on engineering intuition and familiarity with existing materials. Particularly in industries with less prescriptive standards or specifications \cite{prabhu_favoring_2021,karandikar_approach_1992}.  Even with Ashby's systematic approach to material selection, the process is non-trivial and can still leave designers with uncertainty as to how well a candidate material will perform in reality \cite{ashby_materials_2011,ashby_selection_2004,doi:10.1179/mst.1989.5.6.517}. Data and knowledge are essential, without which, limited exploration of alternative or innovative options can occur leading to sub-optimal designs \cite{shiau_optimal_2009,hazelrigg_irrationality_1997}. While established methods like Ashby diagrams can guide designers and encourage them to consider a wider range of possibilities, material databases \cite{ullah_investigation_2020} cannot often account for the ever-growing universe of materials and broadening design considerations outlined above.

The field of material science is currently undergoing a significant transformation due to the advent of artificial intelligence (AI). Traditional, labor-intensive methods of materials discovery are being replaced with automated, parallel, and iterative processes, accelerating the discovery and development of new materials on an unprecedented scale~\cite{gomes2019artificial}. Noteworthy research in this field includes the GNoME project, which claims to have discovered millions of theoretically possible materials using graph neural networks (GNNs) trained on expansive datasets~\cite{merchant2023scaling}. IBM Research has also leveraged AI to pioneer novel battery chemistries that eschew reliance on heavy metals such as cobalt and nickel~\cite{chen2024accelerating}. 

LLMs have been trained with domain-specific materials knowledge to assess their effectiveness in navigating complex topics. Recent studies exploring the readiness of LLMs, specifically those trained with domain-specific data (known as MatSci-LLMs)~\cite{zaki2024mascqa}, suggest that domain-adaptation and task-specific prompting strategies are necessary to extract the desired output from these models. In addition, multimodel datasets require fine-tuning to yield sensible outputs and provide useful insights for deeper material science research~\cite{miret2024llms}.

The rapid pace of material discovery, driven by various AI technologies, presents a challenging question: how can a non-materials expert navigate the ever-growing array of potential materials to identify the best candidates for specific design requirements? The application of LLMs to material selection tasks has emerged as a logical and increasingly crucial area of investigation. This is especially relevant for newly discovered materials, which are relatively unexplored in terms of scalable feasibility and broader property attributes. While previous studies have evaluated the efficacy of LLMs for providing insights into deep, domain-specific knowledge, the evaluation of material selection for component design remains an open area for investigation. 

Even with full visibility of the known material universe, the task of material selection during the conceptual design stage is not without its challenges. For instance, the process can often be subjective, potentially overlooking promising new materials simply because designers are unfamiliar with them \cite{singh_subjective_2022,leontiev_how_2022}. Uncertainty regarding the performance of novel materials can further hinder their adoption. Additionally, manufacturing innovations like additive manufacturing or meta-materials \cite{zadpoor_design_2023, liu_metamaterials_2015, govt_polytechnic_college_review_2015} are allowing previously unfeasible materials to now become viable options \cite{jelinek_design_2015}.  This highlights the need for a data-driven approach to material selection, one that can objectively evaluate a broader range of options while considering the complex interplay of design requirements and provide insights when selecting a particular option \cite{dong_survey_2017}.
Large Language Models (LLMs) offer a powerful new tool for material selection. By learning from vast datasets of past design experiences and material properties, LLMs can provide valuable insights that would otherwise require extensive research or experimentation.

\subsection{Evaluating Large Language Models}
Language models (LMs) are computational models that have the ability to understand and generate human language~\cite{kombrink_recurrent_2011,gao_introduction_2004,devlin_bert_2019}. LMs have the transformative ability to predict the likelihood of word sequences or generate new text based on a given input.

LLMs \cite{kasneci_chatgpt_2023,chen_evaluating_2021,zhao_survey_2023} are advanced language models with large parameter sizes and exceptional learning capabilities. The core module behind many LLMs such as GPT-3 \cite{noauthor_gpt-3_nodate} and GPT-4 \cite{openai_gpt-4_2024} is the self-attention module in Transformer \cite{vaswani2017attention} that serves as the fundamental building block for language modeling tasks. Transformers have revolutionized the field of NLP with their ability to handle sequential data efficiently, allowing
for parallelization and capturing long-range dependencies in text. One key feature of LLMs is in-context learning, where the model is trained to generate text based on a given context or prompt~\cite{brown_language_2020}. This enables LLMs to generate more coherent and contextually relevant responses, making them suitable for interactive and conversational applications.

One common approach to interacting with LLMs is prompt engineering, where users design and provide specific prompt texts to guide LLMs in generating desired responses or performing specific tasks~\cite{white_prompt_2023,zhou_large_2023,clavie_large_2023}. This is widely adopted in existing evaluation efforts. People can also participate in question-and-answer interactions, where they pose questions to the model and receive answers, or engage in dialogue interactions, having natural language conversations with LLM~\cite{jansson_online_2021}.

Assessing the performance of the model is an essential step in evaluating the model. Due to the extensive training data for LLMs, it might not even be feasible to evaluate deep learning models. Thus, evaluation on a static validation set has long been the standard choice for deep learning models \cite{chang2023survey}. For example, computer vision models take advantage of static test sets such as ImageNet \cite{deng_imagenet_2009} and MS COCO \cite{lin_microsoft_2015} for evaluation. LLMs also use GLUE \cite{wang2019glue} or SuperGLUE \cite{wang_superglue_2020} as common test sets. As LLMs are becoming more popular with even poorer interpretability, existing evaluation protocols may not be enough to thoroughly evaluate the true capabilities of LLMs. 

Automated evaluation is a common and perhaps the most popular evaluation method that typically uses standard metrics and evaluation tools to evaluate model performance. Compared to human
evaluation, automatic evaluation does not require intensive human participation, which not only saves time, but also reduces the impact of subjective factors of humans and makes the evaluation process
more standardized \cite{chang2023survey}. 

LLM-EVAL \cite{lin_llm-eval_2023} is a benchmark that is a unified multidimensional automatic evaluation method for open-domain conversations with large language models (LLM). PandaLM \cite{wang_pandalm_2023} is trained to distinguish the superior model given several LLMs. Jain et al. \cite{jain_bring_2023} enabled a more efficient form of evaluating models in real-world deployment by eliminating the need for laborious labeling of new data. Automatic evaluation is also done using standard benchmarks such as MMLU \cite{hendrycks_measuring_2021}, HELM \cite{liang_holistic_2023}, C-Eval \cite{huang_c-eval_2023}, AGIEval \cite{zhong_agieval_2023}, AlpacaFarm \cite{dubois_alpacafarm_2024}, Chatbot Arena \cite{chiang_chatbot_2024}, etc.

The increasingly strengthened capabilities of LLMs have gone beyond standard evaluation metrics on general natural language tasks. Therefore, human evaluation becomes a natural choice in some non-standard cases where automatic evaluation is not suitable. For example, in open-generation tasks where embedded similarity metrics (such as BERTScore) are not sufficient, human evaluation is more reliable \cite{novikova_why_2017}. Human evaluation is a way to evaluate the quality and accuracy of model-generated results through human participation. Compared to automatic evaluation, manual evaluation is closer to the actual application scenario and can provide more comprehensive and accurate feedback. In
manual LLM evaluation, evaluators (such as experts, researchers, or ordinary users) are usually invited to evaluate the results generated by the model. Bubeck et al. \cite{bubeck_sparks_2023}, Bang et al. \cite{bang_multitask_2023}, Liang et al. \cite{liang_holistic_2023}, Ziems et al. \cite{ziems_can_2024}, all manually evaluated LLMs compared to some experts or humans. Although a high variance can be observed in human evaluations attributed to cultural and individual differences \cite{peng_validity_nodate}, from the work of Tjuatja et al. \cite{tjuatja_llms_2024}, they  highlight the pitfalls of using LLMs as human proxies, and underscore the need for finer-grained characterizations of model behavior. This behavior is further backed up by the work of Hopkins et al. \cite{hopkins_can_2023} - LLMs struggle to induce reasonable distributions over generated elements, suggesting that practitioners should more carefully consider the semantics and methodologies of sampling from LLMs.

In this work, we also perform a human evaluation of LLMs and prompt engineering methods by collecting a large dataset of expert-selected scores for a set of materials in specific design scenarios.

\subsection{Automating Material Selection with Machine Learning and Large Language Models}
The process of selecting materials for engineering applications is an interdisciplinary task. It involves an understanding of material properties, structure, and their applications in the context of design requirements. Traditionally, it has relied heavily on the expertise of engineers and scientists and has led to the emergence of a distinct discipline in its own right. Over the years, research has explored the automation of this process, targeting the selection of materials for specific types of objects (e.g., nozzles, beams) and design functions (e.g., heat transfer and storage), often framing material selection as an optimization challenge~\cite{keiser2022material, odum2022numerical,sirisalee2004multi, somkuwar2010materials, chandrasekhar2022integrating, 10.1115/DETC2014-34280, bi2017energy, dehghan2007novel}.

While initial efforts to automate material selection have primarily utilized numerical methods and traditional optimization techniques, there has been a shift towards incorporating machine learning, particularly neural networks, into this process. These approaches, however, often fall short in ranking the suitability of selected materials, limiting their practical utility \cite{mamoonapplication}. A notable exception is the work by Zhou \textit{et al.}, which integrates a two-layer neural network with a genetic algorithm to select sustainable materials, demonstrating the approach through the design of a drink container \cite{zhou2009multi}. Similarly, Chandrasekhar \textit{et al.} employ a variational autoencoder in conjunction with a geometry encoder neural network, allowing for the simultaneous optimization of both the material and the geometry of a beam structure \cite{chandrasekhar2022integrating}. More recent work has leveraged large collections of CAD data with material labels and used graph neural networks to rank the most appropriate materials with a manufacturing- and class-agnostic method~\cite{bian2022material, bian2024hg}.

Advancements in the natural language processing field around transformer-based architectures, coupled with training models on larger and larger datasets, have resulted in several pre-trained Large Language Models (LLMs) capable of mimicking human language, which appear to exhibit emergent reasoning capabilities \cite{brown2020language, vaswani2017attention, wei2022emergent, openai2023gpt4, touvron2023llama, driess2023palm}. In the design engineering domain, recent work has looked at leveraging LLMs to support designers during the conceptual design stage \cite{ma_conceptual_2023, zhu2023generative, jiang2022patent}, as well as for detail design~\cite{picard2023concept, picard2024untrained, meltzer2024s, song2024multi}. Regarding material selection, LLMs have been used to select appropriate materials from an Ashby chart \cite{picard2023concept}, assist with selecting materials for building components \cite{saka2023gpt}, and propose appropriate manufacturing methods \cite{makatura2023large}. LLMs have also been fine-tuned on text extracted from a material textbook to aid in material-related design tasks \cite{buehler2023melm}.

In this work, we further explore the potential for using LLMs to drive material selection. Specifically, we designed a survey to collect a dataset of expert material selection responses, and we conduct a variety of LLM experiments in comparison to that expert data. We investigate a wide variety of prompting approaches and conduct a hyperparameter study over temperature, comparing LLM results to survey responses using both statistical and distance-based metrics.

\section{Methods}

\subsection{Data Collection}
\label{sec:data-collection}
To answer the research questions we first collected a dataset of material preferences by conducting an online survey among professionals with varied experiences in the field of material selection for mechanical design. 

The survey consisted of 4 design cases (\textit{kitchen utensil grip}, \textit{spacecraft component}, \textit{underwater component}, and \textit{safety helmet for sport}) and 4 design criteria (\textit{lightweight}, \textit{heat resistant}, \textit{corrosion resistant} and \textit{high strength}) combined in a full factorial experimental design to produce 16 scenario-based questions. The design cases and criteria were chosen through some pilot studies to maximize the likeness of appropriate and inappropriate material options, and they are also used throughout the LLM experiments. For each question, participants were asked to score a set of nine materials (\textit{steel}, \textit{aluminum}, \textit{titanium}, \textit{glass}, \textit{wood}, \textit{thermoplastic}, \textit{elastomer}, \textit{thermoset}, and \textit{composite}) on a scale from 0 to 10, with 0 being unsatisfactory in the specific application and 10 being an excellent choice. These material categories were chosen to cover a wide range of design use cases, to balance high-level and low-level material categories, and to limit the survey length.  

The survey also collected basic demographic information to ensure that participants had the necessary knowledge and background to provide strong preferences for material selection.
After a series of pilot studies to determine the suitability of the designs, criteria, materials, and data input method, the survey was distributed to professionals who have worked as materials scientists, materials engineers, design engineers, or related fields, through the Autodesk Research Community network, where it remained accessible for 30 days.

\begin{figure}
    \centering
    \includegraphics[width=\columnwidth]{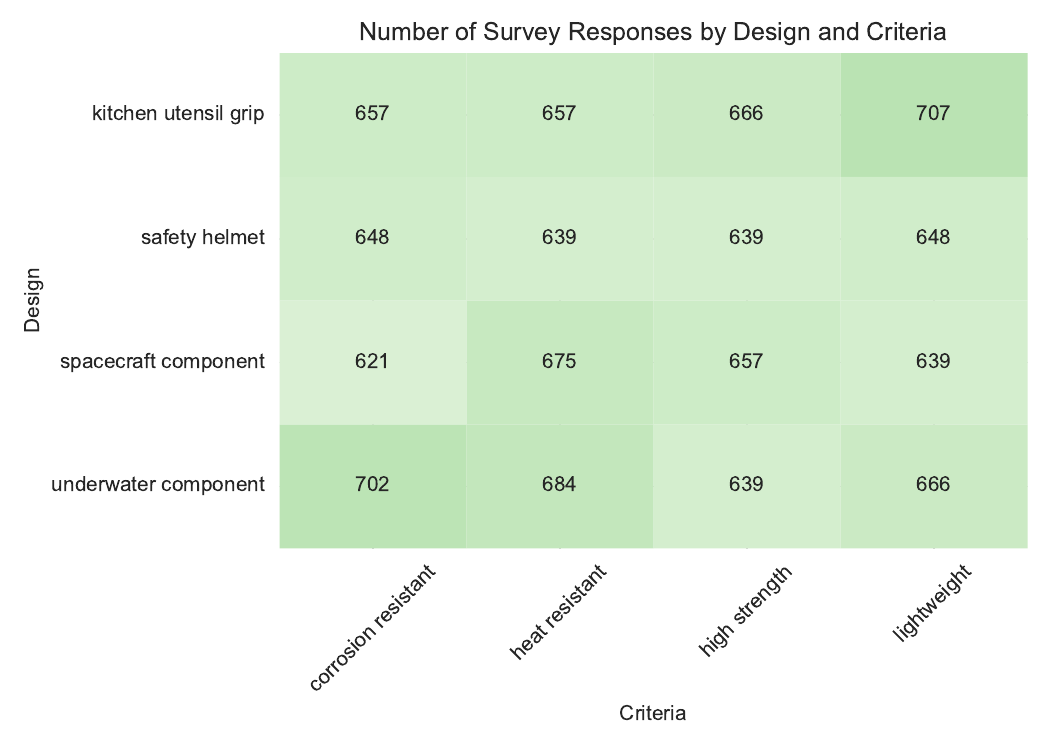}
    \caption{Distribution of the 10,544 survey responses from 136 experts grouped by design and criteria.}
    \label{fig:responses}
\end{figure}

A total of 139 respondents participated in the survey. 136 participants indicated that they had material selection experience, and their responses were kept. As the 16 questions presented to the participants were randomized and independent, we kept all answers even though participants might not have completed every question in the survey. This resulted in a total of 10,544 survey responses across the 9 material categories. The distribution of responses grouped by design and criteria is shown in Figure~\ref{fig:responses}.

\subsection{LLMs used for evaluation}
We perform experiments on three leading models, including one open-source model (Mixtral), one closed-source model (GPT4), and one open-source model fine-tuned on relevant work (MechGPT). The relative strengths of these models and more justification behind their selection is articulated below. 
\begin{enumerate}

\item The closed-source model from OpenAI GPT4 (\texttt{gpt-4-0125-preview}) is chosen because of its high performance on existing benchmarks, including other design tasks \cite{2023GPT4VisionSC, shaham2023zeroscrolls, picard2023concept}. 

\item Mixtral, the open-source model from Mistal AI (\texttt{mixtral-8x7b-instruct-v0.1}) is chosen because of its open-source weights and performance comparable to GPT-3.5~\cite{chiang2024chatbot, jiang2024mixtral}. 

\item Lastly, MechGPT, a recently released fine-tuned \texttt{OpenOrca-Platypus2-13B} model, is chosen as it was fine-tuned specifically on a corpus of mechanics of materials and materials modeling, which might aid in a material selection task for product design~\cite{10.1115/1.4063843, hunterlee2023orcaplaty1}.
\end{enumerate}

We use the default parameters for each of the models as defined in their documentation. We use temperatures of 0.1, 0.75, and 0.4 for GPT4, Mixtral, and MechGPT respectively for each of the experiments described in the following sections (except for the temperature experiment). 

\subsection{LLM Experiments}
To answer our second research question around viable methods to steer LLMs towards expert solutions for a material selection task, we evaluate a series of prompt-engineering methods popular in the natural language processing domain. We also report the effect of temperature as a hyper-parameter for the material selection task. 

For all experiments, we begin the prompt with a preamble, providing some context to the model and introducing each of the four designs and four criteria being tested (substituting \textit{\{design\}} and \textit{\{criterion\}} respectively), as follows:

\begin{customquote} [frametitle={Preamble}]
You are a material science and design engineer expert.

You are tasked with designing a \textit{\{design\}}. The design should be \textit{\{criterion\}}.

\end{customquote}

\subsubsection{Zero-shot}
Research has shown that LLMs exhibit zero-shot reasoning capabilities for arithmetic and other logical reasoning tasks~\cite{kojima2023large}.
Thus, as a baseline, we test a zero-shot approach to material selection, where, for each of the nine materials and 16 design scenarios, the model is prompted to report a value from 0 to 10 to describe the applicability of the material in that specific design scenario. To the preamble described in the previous section, we append the following text, and pass the information to the models:

\begin{customquote} [frametitle={Zero-shot Prompt}]
How well do you think \textit{\{material\}} would perform in this application? Answer on a scale of 1-10, where 0 is `unsatisfactory', 5 is `acceptable', and 10 is `excellent', with just the number and no other words.
\end{customquote}

In this prompt, \textit{\{material\}} is substituted for each of the nine materials being tested, resulting in a total of 144 prompts.The wording of the zero-shot prompt was chosen to match the questions posed to the survey participants.

\subsubsection{Few-shot}
\label{sec:few-shot}
Few-shot prompting involves providing the model with one or more demonstrative examples, with the intent of giving some task-specific context to the model, which has been shown to surpass a zero-shot approach across a range of benchmarks~\cite{brown2020language}.

In the context of our material selection task, we take a two-shot approach, where we include in the prompt two examples not present in our survey formulation previously described in Section~\ref{sec:data-collection} and shown in Figure~\ref{fig:overview}. The two examples are a `bicycle frame grip' that needs to be `impact resistant', and a `medical implant grip' that needs to be `durable'. We assemble the prompt for this method by appending to the preamble the following text, where \texttt{\{few_shot\}} is replaced by the aforementioned two examples (the full text of which can be found in Appendix~\ref{app:few-shot}:

\begin{customquote} [frametitle={Few-shot Prompt}]
Below are two examples of how materials would perform from 1-10 given a design and a criterion:

\textit{\{few_shot\}}

How well do you think \textit{\{material\}} would perform in this application? Answer on a scale of 1-10, where 0 is 'unsatisfactory', 5 is 'acceptable', and 10 is 'excellent', with just the number and no other words.
\end{customquote}

\subsubsection{Parallel Agents}
While the zero-shot and few-shot methods ask the model to provide a score for each of the nine materials separately (in serial), another method is to ask the model for the scores of all nine materials simultaneously (in parallel). Asking the model for all nine scores in the same prompt reduces the number of prompts from 144 to just 16. Thus, this method has the benefit of increasing scalability to a much larger number of material categories. However, parallelizing requires more post-processing of the output, and introduces more opportunities for model hallucination, given the higher complexity of the question.

Initial experiments showed that at least one of the models we evaluated required more detailed instructions to give scores for all nine materials in a consistent manner. The additional clarifications can be seen in the following prompt, where again, we start with the preamble and append the following text:

\begin{customquote} [frametitle={Parallel Prompt}]
For each of the following materials, how well do you think they would perform in this application? Answer on a scale of 1-10, where 0 is `unsatisfactory', 5 is `acceptable', and 10 is `excellent', just with the integers separated by commas, and no other words or explanation. Be concise and answer for all 9 materials.

Materials:

\textit{\{materials\}}

Answers:
\end{customquote}

\subsubsection{Chain-of-thought}
\label{sec:chain-of-thought}
Prior work has shown that generating a chain of thought, or intermediate reasoning steps, significantly improved the ability of LLMs to perform different reasoning tasks~\cite{wei2022chain}. This method has been expanded to tree-of-thoughts (to include multiple chains) \cite{yao2024tree}, and generalized in methods that allow the model to reflect on the answers, provide more explanations, or correct itself~\cite{shinn2023reflexion, white2023prompt}. 

In this paper, we implement a two-step chain-of-thought, where we first ask the model to describe how the material would perform in the specific design scenario, and include that reasoning in the final question where we ask the model to give a score. To generate the reasoning, we append the following text to the preamble:

\begin{customquote} [frametitle={Chain-of-Thought Prompt 1}]
How well do you think \textit{\{material\}} would perform in this application?
\end{customquote}

Then, the reasoning is included in the following prompt as \texttt{\{reasoning\}} (which is again appended to the preamble):

\begin{customquote} [frametitle={Chain-of-Thought Prompt 2}]
How well do you think \textit{\{material\}} would perform in this application? Below is some reasoning that you can follow:

\textit{\{reasoning\}}

Answer on a scale of 1-10, where 0 is 'unsatisfactory', 5 is 'acceptable', and 10 is 'excellent', with just the number and no other words.
\end{customquote}

\subsubsection{Temperature}
\label{sec:temp}
Temperature is an LLM hyperparameter that controls the sampling probability when selecting the next token. The three LLMs tested all have different guidance and defaults when choosing the temperature parameter for generation. For all prompt engineering experiments described previously, we use the default temperature values of 0.1 for GPT-4, 0.75 for Mixtral, and 0.4 for MechGPT. 

In prior literature there are mixed results regarding the importance of this hyperparameter for question and answering tasks or more design-focused tasks~\cite{renze2024effect, ma2023conceptual}.
Thus, we add to the discourse in this work by including this parameter as a controlled variable, and report zero-shot results by sweeping the temperature from 0 to 1 for GPT-4 and Mixtral, and from 0.1 to 1 for MechGPT (as it does not support a temperature of 0).

\subsection{Evaluation}
To answer the first research question, we seek to quantify the amount of bias that LLMs have toward specific materials with two metrics the z-score and the mean distance to the survey data.

In statistics, the z-score, or standard score, measures the number of standard deviations by which a value $x$ is above or below the mean $\mu$ of a distribution, whose standard deviation is $\sigma$:

\begin{equation}
z = \frac{x - \mu}{\sigma} 
\end{equation}

We compute the z-score for each LLM-generated score relative to the survey data grouped by design, criteria, and material. We then take the mean of the z-scores, as an average measure of how far above or below from the mean the LLMs score. The z-score is valid between -1 and 1, and a z-score closer to 0 indicates that, on average, the LLMs' responses are more aligned with the survey data.

While the z-score tends to be used in the context of normal curves, we note that some of the survey data aggregated by material are not normal, and exhibit bi-modal distributions. To mitigate possible negative effects of this, we also report the mean absolute error (MAE) of the generated values to the survey data. By treating the problem as a regression task, the MAE can give a sense of the average deviation between the human rankings and the model predictions. Unlike the z-score, the MAE does not consider whether the deviation is positive or negative. To calculate the MAE, we first group the results by design scenarios and material, then, for each value generated by the LLMs, we compute the mean distance to each survey data in the relative group as 

\begin{equation}
    \text{MAE} = \frac{1}{n} \sum_{i=1}^{n} |y_i - \hat{y}_i|
\end{equation}

where $n$ is the number of survey responses for the specific design scenario and material, $y_i$ is the $i$th expert score, $\hat{y}_i$ is the $i$th predicted score by the LLM, $|y_i - \hat{y}_i|$ represents the absolute error of the $i$th prediction. The MAE is valid from 0 to 10, and a value closer to 0 indicates that, on average, the LLMs' responses are closer to the survey participants' responses.

\section{Results and Discussion}

The mean z-scores and MAE results for all experiments and models are shown in Table~\ref{tab:z-scores} and~\ref{tab:distance} respectively.

\begin{figure}[]
    \centering
    \includegraphics[width=\linewidth]{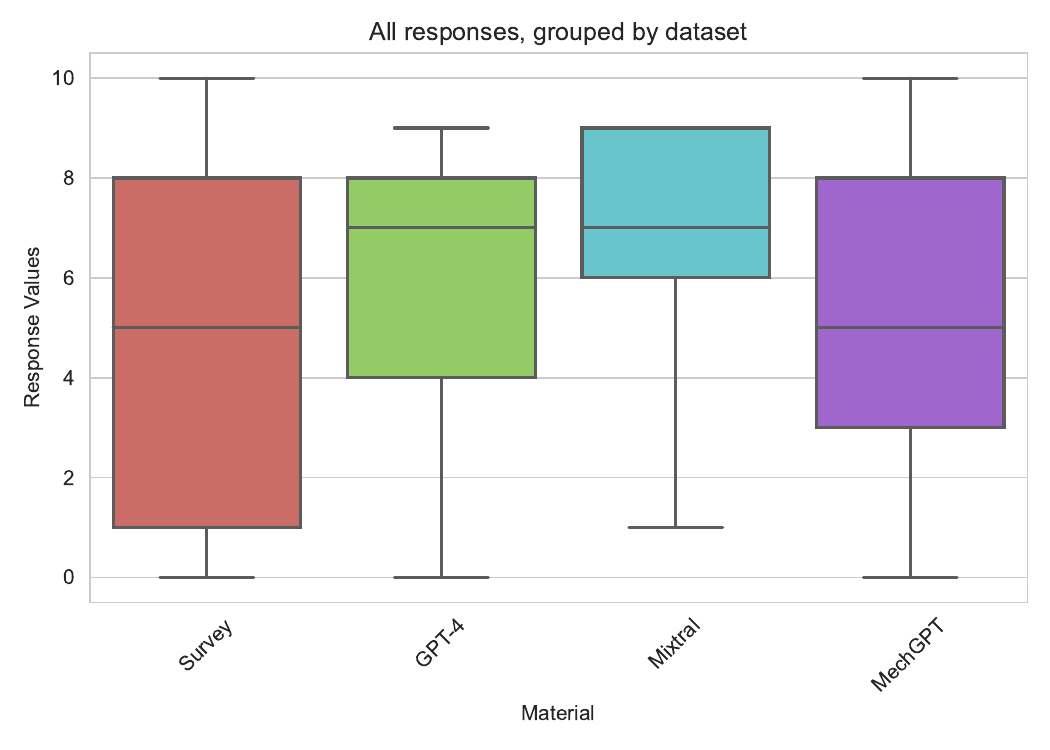}
    \caption{Aggregate survey and zero-shot LLM responses, showing the full range and quartiles across all designs, criteria, and materials.}
    \label{fig:responses-by-dataset}
\end{figure}

\begin{figure*}[ht!]
\centering
\includegraphics[width=\linewidth]{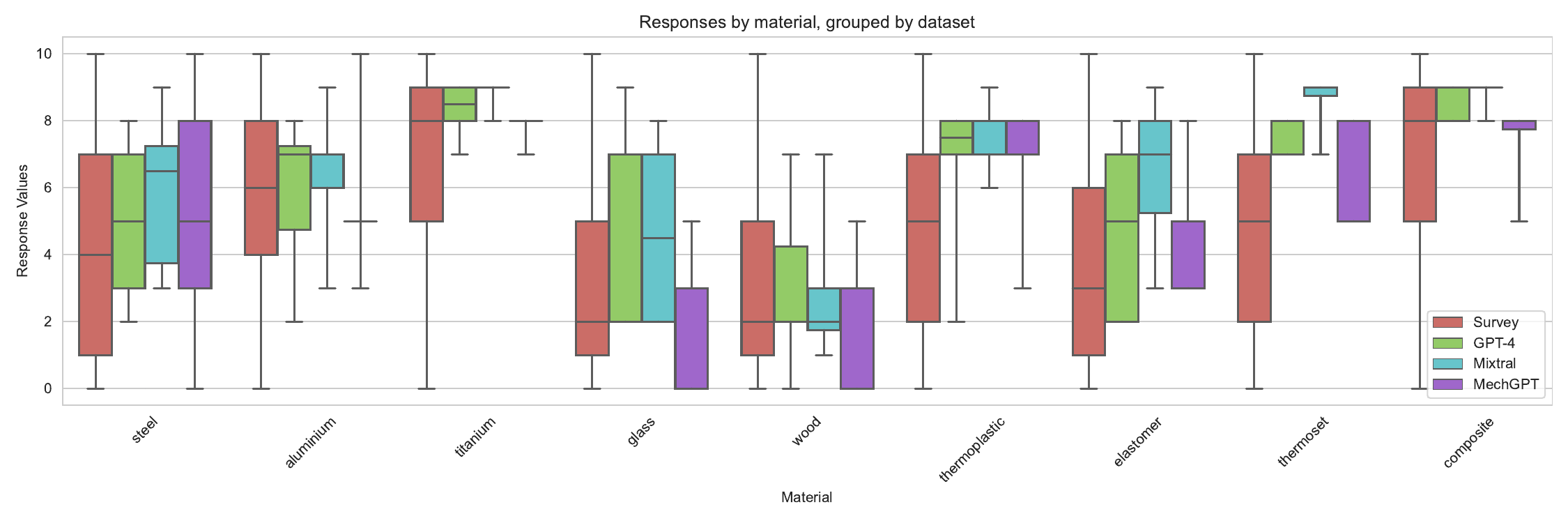}
\caption{\label{fig:aggregate}
Aggregate survey and zero-shot LLM results grouped by material.}
\end{figure*}

\begin{figure*}[ht!]
    \centering
    \includegraphics[width=\linewidth]{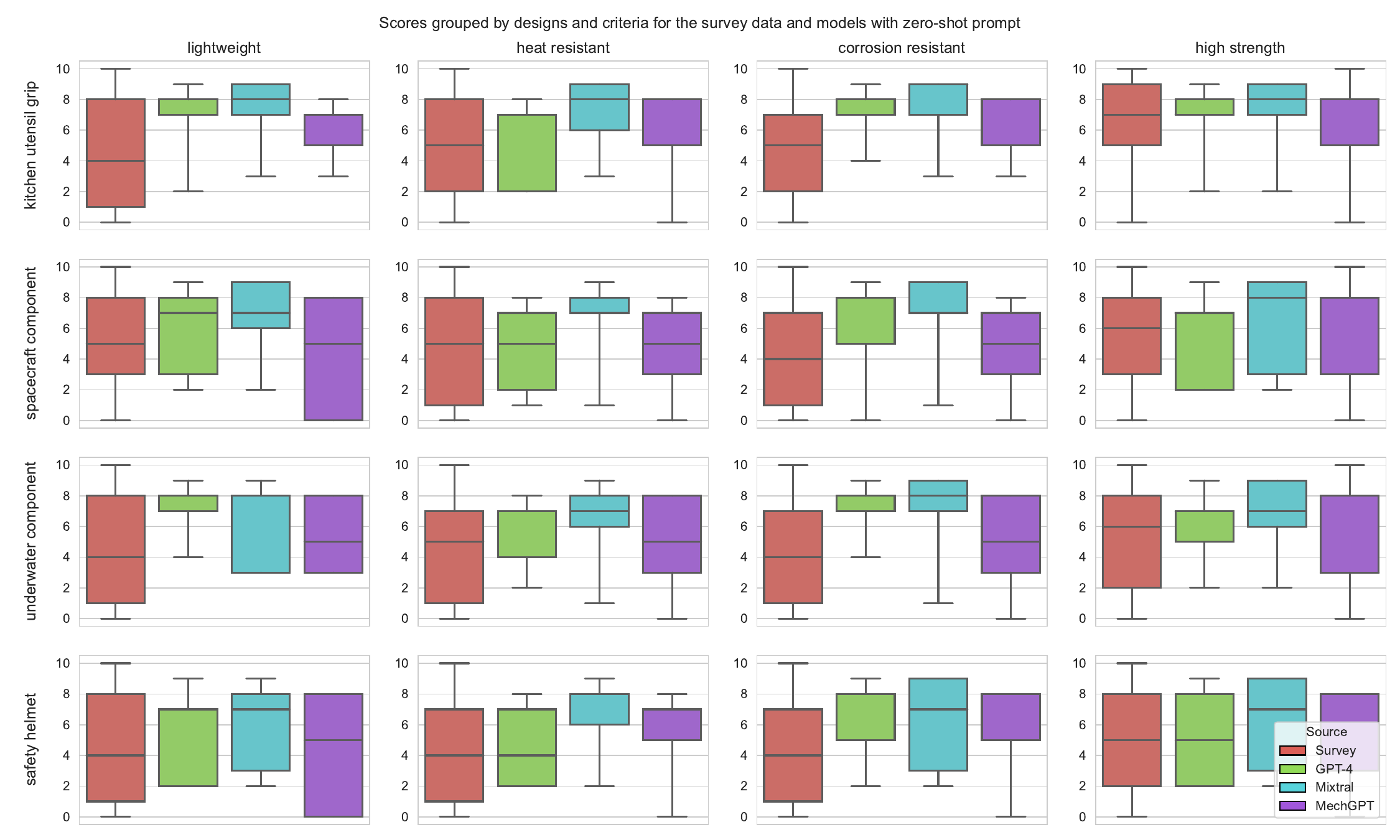}
    \caption{\label{fig:aggregate-design-criteria}
    Aggregate survey and zero-shot LLM results grouped by design (rows) and criteria (columns).}
\end{figure*}

The survey and zero-shot LLM responses are aggregated in Figure~\ref{fig:responses-by-dataset}. The zero-shot results are further broken down for each material in Figure~\ref{fig:aggregate}, and for each design and criteria in Figure~\ref{fig:aggregate-design-criteria}. Table~\ref{tab:mean-error} shows the three largest and negative mean errors by GPT-4.

\begin{figure}[b]
\centering
\includegraphics[width=\columnwidth]{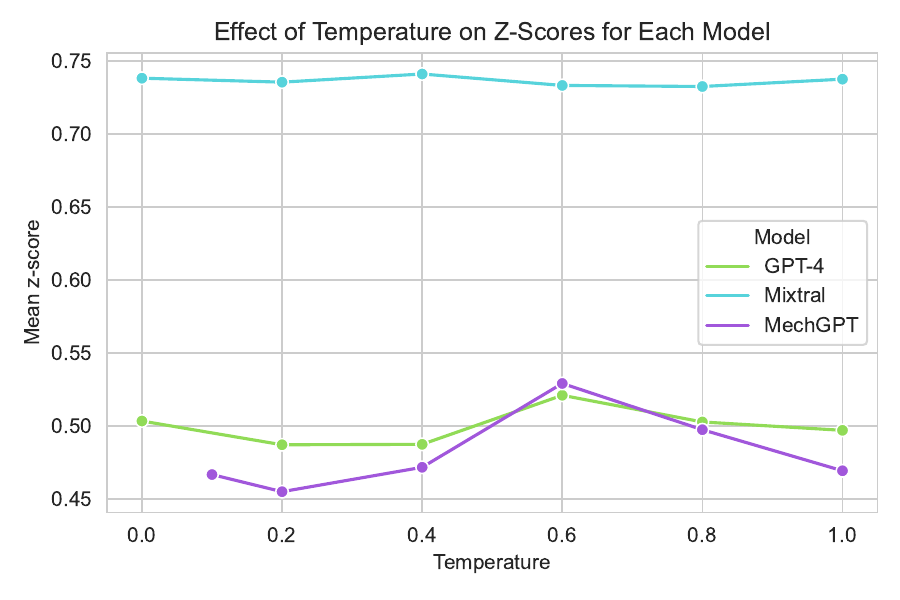}
\caption{\label{fig:temperature-z-score}
The effect of changing the temperature hyperparameter on the z-score for zero-shot generation.}
\end{figure}

\begin{figure}[b]
\centering
\includegraphics[width=\columnwidth]{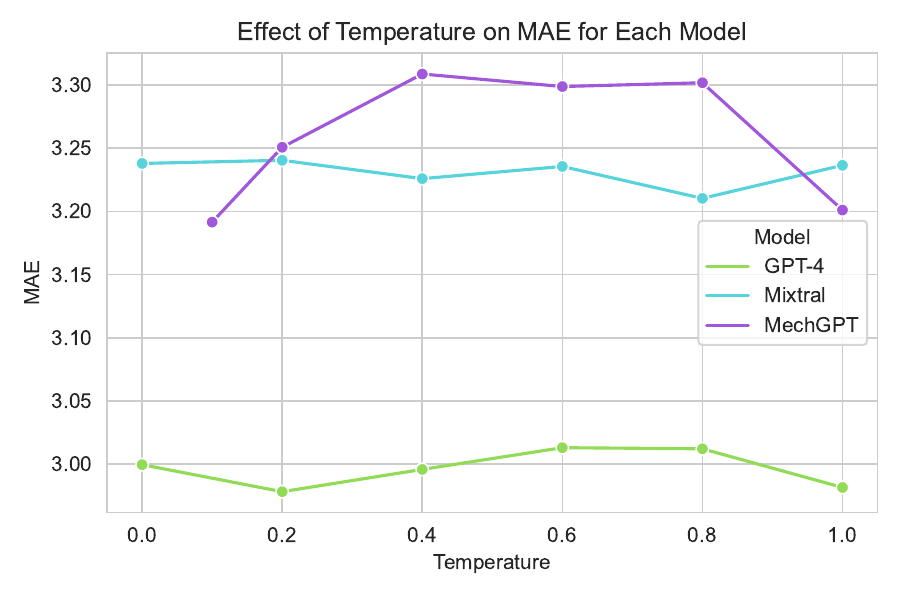}
\caption{\label{fig:temperature-distance}
The effect of changing the temperature hyperparameter on MAE for zero-shot generation.}
\end{figure}

Figures~\ref{fig:temperature-z-score} and \ref{fig:temperature-distance} show the results from the temperature sweep experiments for the zero-shot method across the different LLMs.

\begin{table}[b]
\centering
\caption{Mean z-scores for all experiments across the three models we tested. The z-score measures how far a sample is from the mean of a set, which in this case is the distance of the generated samples to the mean of the survey results. }
\label{tab:z-scores}
\resizebox{\columnwidth}{!}{
\begin{tabular}{@{}lllllll@{}}
\toprule
 & \multicolumn{3}{c}{\textbf{Models}} &  &  \\ \midrule
 & \textbf{GPT-4} & \textbf{Mixtral} & \textbf{MechGPT} &  & \textbf{Mean} \\ \midrule
\textbf{Zero-shot} & \cellcolor[HTML]{B1D68B}0.451 & \cellcolor[HTML]{EAE897}0.722 & \cellcolor[HTML]{77C47F}0.179 &  & \cellcolor[HTML]{D8E293}0.451 \\
\textbf{Few-shot} & \cellcolor[HTML]{B9D98D}0.492 & \cellcolor[HTML]{BBD98D}0.501 & \cellcolor[HTML]{CADE90}0.572 &  & \cellcolor[HTML]{F0EA98}0.522 \\
\textbf{Parallel} & \cellcolor[HTML]{A1D188}0.379 & \cellcolor[HTML]{8ACA83}0.269 & \cellcolor[HTML]{93CD85}-0.312 &  & \cellcolor[HTML]{63BE7B}0.112 \\
\textbf{Chain-of-thought} & \cellcolor[HTML]{FFEF9C}0.821 & \cellcolor[HTML]{F7EC9A}0.787 & \cellcolor[HTML]{63BE7B}0.081 &  & \cellcolor[HTML]{FFEF9C}0.563 \\
\textbf{} &  &  &  &  &  \\
\textbf{Mean} & \cellcolor[HTML]{F2EB99}0.536 & \cellcolor[HTML]{FFEF9C}0.570 & \cellcolor[HTML]{63BE7B}0.130 &  &  \\ \bottomrule
\end{tabular}%
}
\end{table}

\begin{table}[b]
\centering
\caption{Mean absolute errors (MAE) to survey data for all experiments across the three models we tested. A lower MAE indicates that the model is generating, on average, values closer to the actual survey responses.}
\label{tab:distance}
\resizebox{\columnwidth}{!}{
\begin{tabular}{@{}lllllll@{}}
\toprule
                          & \multicolumn{3}{c}{\textbf{Models}}                                                           &  &                               \\ \midrule
                          & \textbf{GPT-4}                & \textbf{Mixtral}              & \textbf{MechGPT}              &  & \textbf{Mean}                 \\ \midrule
\textbf{Zero-shot}        & \cellcolor[HTML]{79C57F}2.949 & \cellcolor[HTML]{C3DC8F}3.243 & \cellcolor[HTML]{8ACA83}3.015 &  & \cellcolor[HTML]{6EC17D}3.069 \\
\textbf{Few-shot}         & \cellcolor[HTML]{A5D288}3.121 & \cellcolor[HTML]{C1DB8F}3.234 & \cellcolor[HTML]{C9DE90}3.265 &  & \cellcolor[HTML]{AFD68B}3.207 \\
\textbf{Parallel}         & \cellcolor[HTML]{70C27D}2.911 & \cellcolor[HTML]{63BE7B}2.859 & \cellcolor[HTML]{E2E696}3.365 &  & \cellcolor[HTML]{63BE7B}3.045 \\
\textbf{Chain-of-thought} & \cellcolor[HTML]{FFEF9C}3.477 & \cellcolor[HTML]{D5E293}3.314 & \cellcolor[HTML]{DAE394}3.332 &  & \cellcolor[HTML]{FFEF9C}3.374 \\
\textbf{}                 &                               &                               &                               &  &                               \\
\textbf{Mean}             & \cellcolor[HTML]{63BE7B}3.115 & \cellcolor[HTML]{9CD087}3.163 & \cellcolor[HTML]{FFEF9C}3.244 &  &                               \\ \bottomrule
\end{tabular}
}
\end{table}

\subsection{Aggregate evaluation of LLMs against expert results}

Focusing on the survey results shown in Figure~\ref{fig:aggregate} we see that the minimum and maximum values selected by experts vary from 0 to 10 for all materials across the design scenarios. Also, each material has a large variance of appropriateness as seen in the quartiles of the box plot (aluminum has the smallest standard deviation of 2.85, while steel has the largest standard deviation of 3.23). This is expected, as the 16 design scenarios in the survey were chosen to maximize the appropriateness of the materials. Nonetheless, looking at the median for each material, we see that some materials are considered more appropriate across all 16 design scenarios than others. For example, titanium and composite materials have the highest median of 8. On the other hand, glass and wood have the lowest medians of 2. Of the plastics, thermoplastic and thermoset score higher than elastomers, on average. While aluminum scores higher than steel, with a median score of 6 and 4 respectively. 

Looking at all the LLM results grouped in Figure~\ref{fig:responses-by-dataset}, we see that, differently from the survey responses, GPT-4 and Mixtral do not span quite the full breadth of the 0-10 scale. Moreover, the overall variance of the LLMs is smaller than the experts, possibly indicating expert results are highly variable due to the differences in people's experiences, familiarity with different materials and, possibly, disagreements between experts driven by other design assumptions not specified in the design scenario. On the other hand, LLM responses are less diverse, possibly due to the RLHF training driving more homogeneous responses across different design scenarios. 

In Figure~\ref{fig:aggregate}, we can see similar trends between experts and LLMs regarding the relative appropriateness of the materials. For all models, titanium and composites score the highest, and glass and wood score the lowest, which matches the survey results. However, there is a noticeable difference in the variance of the LLM responses compared to the expert responses for some materials. Mixtral gives titanium a 9 regardless of the design scenario (SD=0.250), MechGPT also most often scores titanium as 8 (SD=0.342), and GPT-4 most often gives thermosets a 7 (SD=0.479). In aggregate, the expert results have a standard deviation of 3.346, while GPT-4, Mixtral, and MechGPT have standard deviations of 2.486, 2.488, and 2.933, respectively. This shows the three LLMs tested are less biased by the design and criterion provided in the prompt, compared to experts, and are instead more biased by the material itself, and possibly prior information about the materials that was found in the training corpus. 

In general, models tend to score all materials well above experts as shown in Tables~\ref{tab:z-scores} and~\ref{tab:distance}, where all MAEs are around 3, on the 0 to 10 scale. Only MechGPT when used in combination with the parallel prompt resulted in a negative z-score, indicating that the mean is below that of the experts (Table~\ref{tab:z-scores}). This is driven by MechGPT scoring many materials as 0, when prompted in parallel. When using the zero-shot prompt, Figure~\ref{fig:aggregate} also shows that LLMs particularly score steel and plastics (thermoplastic, elastomer, and thermoset) higher than experts, compared to the other materials.

It is difficult to draw conclusions about which model performs better at this task, as the two metrics we evaluate show different trends. In Table~\ref{tab:z-scores}, showing the mean z-scores for each model, the best-performing model is MechGPT, but the negative value of the parallel prompt drives down its mean. However, MechGPT still outperforms Mixtral and GPT-4 in the zero-shot and chain-of-thought prompts by a wide margin, and performs similarly to the other models when using few-shot and parallel prompts. However, using the mean distance metric, MechGPT underperforms with a mean distance of 3.244, while Mixtral, on average, generates values 3.163 away from expert values, and GPT-4 is the best-performing model, with a mean distance of 3.155.

\subsection{Zero-shot}
Using a zero-shot approach for the material selection problem results in an MAE of 3.069 and a z-score of 0.451 averaged across all models. MechGPT performs impressively well compared to the other models, with a narrow z-score of 0.179, followed closely by GPT-4 in terms of MAE. However all models, on average, score the appropriateness of materials 3 points above experts, which is a qualitatively large error. For zero-shot prompts, Mixtral is the worst performing model, with a z-score of 0.722, and an MAE of 3.243.

Further breaking down the zero-shot results by design and criteria (Figure~\ref{fig:aggregate-design-criteria}) highlights some interesting trends. The minimum and maximum values of the survey results go from 0 to 10, which is once again expected, as some materials will be very appropriate for a specific design scenario, while others will be not appropriate at all. The lower and upper quartile values of the survey results also tend to span a large amount of the scale, except for the \textit{high strength} \textit{kitchen utensil grip}, which appear to have a smaller variance than other scenarios. The LLMs however, do not exhibit this broad behavior, both in terms of minimum and maximum values, as well as in terms of how much the quartiles span. For example, all materials seem to be appropriate for \textit{lightweight kitchen utensil grip}, and \textit{corrosion resistant kitchen utensil grip}, which is unexpected and does not reflect the expert responses.

While aggregating results by model, materials, and design scenarios can help conclude general trends, we can also isolate the most extreme errors in the data. The mean largest and smallest errors by GPT-4 using the zero-shot prompt are shown in Table~\ref{tab:mean-error}. GPT-4 has a positive mean error of 6.21 for \textit{elastomer} in the context of a \textit{lightweight} \textit{kitchen utensil grip}. On the opposite end, the model indicates that \textit{steel} might not be appropriate for a \textit{lightweight safety helmet}, scoring it on average -4.58 points away from the experts. Given the lack of transparency and explainability in the LLM's outputs, it is difficult to draw conclusions about the reasons for these discrepancies. One explanation is that the model might not have seen in its training data examples of lightweight steel helmets.

\subsection{Few-shot}
Considering the MAE of all models, few-shot prompting does not seem to improve on zero-shot prompting, increasing the MAE from 3.069 to 3.207. The z-score also increases from 0.451 for zero-shot, to 0.522 for few-shot prompts. Mixtral is the only model to improve the z-score with few-shot prompting (from 0.722 to 0.501), with a marginal improvement in MAE. This loss in performance could be due to the two examples chosen (found in Appendix~\ref{app:few-shot}) not being sufficient for the model to learn in-context, or more examples being required for this task.

\subsection{Parallel Agents}
The parallel agent prompting method achieves the best MAE of 3.045 averaged across the three models. Regarding z-scores, the parallel prompt helped both GPT-4 and Mixtral achieve their best results (0.379 and 0.269 respectively). Notably, this was the only method where a model, MechGPT, had a negative mean z-score of -0.312, driven by the presence of many 0's in the generated answers. However,  the driving factor behind the many 0's in MechGPT's answers is unclear: it could be a failure mode of the model, or the parallel prompting might have resulted in more extreme classification.  

\begin{table}[t!]
\centering
\resizebox{\columnwidth}{!}{
\begin{tabular}{l l l r}
\toprule
\textbf{Design}                & \textbf{Criteria}            & \textbf{Material}    & \textbf{Mean Error} \\
\midrule
Kitchen utensil grip           & Lightweight                  & Elastomer            & 6.21                \\
Underwater component           & Lightweight                  & Wood                 & 5.47                \\
Underwater component           & Corrosion resistant          & Thermoplastic        & 5.37                \\
\multicolumn{4}{c}{\textit{...}} \\
Kitchen utensil grip           & Heat resistant               & Aluminium            & -4.21               \\
Kitchen utensil grip           & High strength                & Glass                & -4.47               \\
Safety helmet                  & Lightweight                  & Steel                & -4.58               \\\bottomrule
\end{tabular}
}
\caption{Largest positive and negative mean errors by GPT-4 using the zero-shot prompt.}
\label{tab:mean-error}
\end{table}

In our testing, we found that parallelizing material generation greatly reduces computational times (from 3 minutes to 35 seconds for GPT-4), since the total number of prompts falls from 144 to 16. This has the potential to greatly improve the scalability of the material search, where more than 9 material categories could be more cheaply evaluated in a single prompt. However, the parallel generation required a lot of manual error-checking to post-process the output (especially for MechGPT), since the models did not follow the prompt to the letter every time. This limitation could potentially be addressed by combining the parallel agents prompting with some few-shot examples.

Overall, our experiments show that parallelizing the prompt and asking the model to evaluate multiple materials at the same time, has both computational and qualitative benefits.

\subsection{Chain-of-Thought}
Using the chain-of-thought approach, as implemented in Section~\ref{sec:chain-of-thought}, did not seem to benefit the material selection task.  GPT-4 and Mixtral achieved the worst results both in terms of z-score (0.821 and 0.787 respectively) and MAE (3.477 and 3.314). On the other hand, using this prompting method, MechGPT achieved the best z-score (0.081), but a high MAE (3.332).  It is possible that MechGPT's fine-tuning is driving the impressive results when using this chain-of-thought prompting method.  

While running the experiment, we found that the chain of thought method greatly increased computational times (from 3 to 57 minutes for GPT4), since we are not limiting the length of the reasoning generation. This could be curbed by limiting the maximum number of tokens generated by the models. Moreover, other implementations of the chain-of-thought method have been developed (such as simply asking the model to `think step by step'~\cite{kojima2023large}), which might yield better results and could be assessed in future work. 

Aside from the apparent loss in performance and increase in computational time, the chain-of-thought approach benefits from increased explainability behind the ultimate results. While the focus of this paper is to compare the raw scores on a 0 to 10 scale to select an appropriate material, it is important to note that in practice, the benefit of the additional reasoning provided by the chain-of-thought approach might greatly benefit a designer hoping to use an LLM for this task.

\subsection{Temperature}
As the three models tested all suggest different temperature values in their documentation, the temperature experiment sought to evaluate the effects that temperature might have on the material selection task. Figures~\ref{fig:temperature-z-score} and Figure~\ref{fig:temperature-distance} show the effect of varying the temperature on the z-scores and MAE respectively. The models do not appear to be affected by temperature when measured using the z-score, with only GPT-4 and MechGPT exhibiting a slight improvement at a temperature of 0.6. When measuring MAE, GPT-4 and Mixtral are unaffected by changing temperature, while MechGPT shows a slight dip at the extremes (0.1 and 1.0). 

These results indicate that the default model values (listed in Section~\ref{sec:temp}) are suitable for all experiments conducted in this work. Furthermore, the material selection task does not seem to be significantly affected by this hyperparameter, which echoes the results found in other work~\cite{renze2024effect}.

\section{Limitations and Future Work}
This work is limited by several factors, each of which implies valuable future work. These areas include explainability, design scenario authenticitiy, prompt generation methodology, and the range of models evaluated.

Most of the experiments presented in this work focus on giving materials a score from 0 to 10, without requiring any additional justification about why the material is appropriate. A notable exception is the chain-of-thought prompt, but even then the reasoning exhibited in the response is not evaluated in this work. This leaves a substantial gap in research around the usefulness and usability of these generated scores when performing a material selection task in practice. In future work, a user study should be conducted to understand how people might interact with LLMs during the material selection process, when providing this information would be most beneficial to the designer, and whether justifying the reasoning could help further inspire the designer towards better material choices. 

The design scenarios used in this study are limited in number and relatively simple. Including multiple conflicting design criteria would make the scenarios more realistic, and increase the applicability of the results. This study also assumes that the expert solutions are the ground truth. However, material selection is driven by careful consideration of conflicting design tradeoffs. Thus, further research should evaluate more complex scenarios and potential LLM biases towards certain design criteria.

The current study was limited to a small set of prompt engineering methods to control the generation process, which could be expanded further. Also, combining prompting methods, such as chain-of-thought with parallel prompting (which performed best), might improve the results and increase the explainability and transparency of the generated responses.

More models could be evaluated which might perform better. Furthermore, while we evaluate a fine-tuned model, MechGPT, it is important to note that the dataset used for fine-tuning may not align perfectly with the task of material selection, as it is primarily focused on failure mechanics of materials. Thus, fine-tuning on a different corpus, or implementing a retrieval augmented generation (RAG) system might improve the results.

\section{Conclusions}
In summary, this work provides insights into the efficacy of large language models (LLMs) in the material selection process, a critical aspect of conceptual design. We identified two notable failure modes: the low variance of recommendations across different design scenarios and the tendency toward the overestimation of material appropriateness. These findings underscore the challenges in adapting current LLMs directly for nuanced tasks like material selection, where diversity in recommendations and accurate assessment of material characteristics are paramount.

Our exploration into the robustness of these models, particularly with respect to temperature, revealed that certain methodologies, such as the parallel prompting method, could enhance the LLMs' performance by mitigating some of the identified failure modes. The parallel prompting method not only demonstrated improved quality of the outputs but could also be leveraged to scale up to a larger number of materials efficiently. 


\section*{Acknowledgment} 

We thank the anonymous survey participants in the Autodesk Research Community for their contribution.




\appendix   

\section{Few-Shot Prompt}
\label{app:few-shot}

The following is the text we use in the few-shot prompt, defined as \textit{\{few_shot\}} in Section~\ref{sec:few-shot}:

\begin{customquote} [frametitle={Few-Shot Prompt Details}]
- Example 1

        You are tasked with designing the grip of a bicycle frame which should be impact resistant.
        
        How well do you think each of the provided materials would perform in this application? (Use a scale of 0-10 where 0 is `unsatisfactory', 5 is `acceptable', and 10 is `excellent').
    
        Steel: 6 
        
        Aluminium: 5 
        
        Titanium: 4
        
        Glass: 2 
        
        Wood: 8 
        
        Thermoplastic: 9 
        
        Elastomer: 9 
        
        Thermoset: 6 
        
        Composite: 7

- Example 2

        You are tasked with designing a medical implant which should be durable. 
        
        How well do you think each of the provided materials would perform in this application? (Use a scale of 0-10 where 0 is `unsatisfactory', 5 is `acceptable', and 10 is `excellent').
        
        Steel: 7 
        
        Aluminium: 2 
        
        Titanium: 9 
        
        Glass: 5 
        
        Wood: 0 
        
        Thermoplastic: 8 
        
        Elastomer: 8 
        
        Thermoset: 7 
        
        Composite: 7
\end{customquote}



\bibliographystyle{asmejour}   

\bibliography{asmejour-sample} 



\end{document}